\documentclass[10pt,twocolumn,letterpaper]{article}

\usepackage{bm}
\usepackage{amsmath}
\usepackage{times}
\usepackage{epsfig}
\usepackage{graphicx}
\usepackage{amsmath}
\usepackage{amssymb}
\usepackage{pifont}
\usepackage{makecell}
\usepackage{bm}
\usepackage{multirow}
\usepackage{booktabs}    
\usepackage{makecell}    
\usepackage{pifont}    
\usepackage{booktabs}   
\usepackage{pifont}     
\usepackage{siunitx}    
\usepackage{array}      
\usepackage{colortbl}
\usepackage[pagenumbers]{iccv} 

\definecolor{iccvblue}{rgb}{0.21,0.49,0.74}
\usepackage[pagebackref,breaklinks,colorlinks,allcolors=iccvblue]{hyperref}

\def\paperID{8560} 
\def\confName{ICCV}
\def\confYear{2025}

\title{ AP-CAP: Advancing High-Quality Data Synthesis for Animal Pose Estimation via a Controllable Image Generation Pipeline}

\author{
Lei Wang\textsuperscript{1,2}, Yujie Zhong\textsuperscript{1†}, Xiaopeng Sun\textsuperscript{1}, Jingchun Cheng\textsuperscript{2},\\ Chengjian Feng\textsuperscript{1}, Qiong Cao, Lin Ma\textsuperscript{1},  Zhaoxin Fan\textsuperscript{2†} \\
\textsuperscript{1}Meituan Inc., \textsuperscript{2}Beihang University \\
\texttt{\{wanglei290, zhongyujie\}@meituan.com, zhaoxinf@buaa.edu.cn}
}

\begin{document}
\maketitle
\renewcommand{\thefootnote}{\fnsymbol{footnote}} 
\setcounter{footnote}{0} 
\footnotetext{\textsuperscript{†}Corresponding authors.}
\begin{abstract}
The task of 2D animal pose estimation plays a crucial role in advancing deep learning applications in animal behavior analysis and ecological research. 
Despite notable progress in some existing approaches, our study reveals that the scarcity of high-quality datasets remains a significant bottleneck, limiting the full potential of current methods. 
To address this challenge, we propose a novel Controllable Image Generation Pipeline for synthesizing animal pose estimation data, termed AP-CAP. Within this pipeline, we introduce a Multi-Modal Animal Image Generation Model capable of producing images with expected poses. 
To enhance the quality and diversity of the generated data, we further propose three innovative strategies: 
(1) Modality-Fusion-Based Animal Image Synthesis Strategy to integrate multi-source appearance representations, 
(2) Pose-Adjustment-Based Animal Image Synthesis Strategy to dynamically capture diverse pose variations, 
and (3) Caption-Enhancement-Based Animal Image Synthesis Strategy to enrich visual semantic understanding. 
Leveraging the proposed model and strategies, we create the MPCH Dataset (Modality-Pose-Caption Hybrid), 
the first hybrid dataset that innovatively combines synthetic and real data, establishing the largest-scale multi-source heterogeneous benchmark repository for animal pose estimation to date. 
Extensive experiments demonstrate the superiority of our method in improving both the performance and generalization capability of animal pose estimators.
\end{abstract}    
\section{Introduction}
\label{sec:intro}
Recognized as a foundational tool for advancing scientific understanding and addressing practical challenges in agriculture, conservation, and environmental science, 2D animal pose estimation focusing on predicting keypoints and thereby inferring the pose from images or videos, is a crucial task with diverse applications, such as animal  monitoring in precision livestock farming and non-invasive tracking and behavioral analysis of animals in  wildlife conservation \cite{straka2024hitchhiker, ye2024superanimal, lauer2022multi}.
\begin{figure}[t]
\begin{center}
\includegraphics[width=\linewidth]{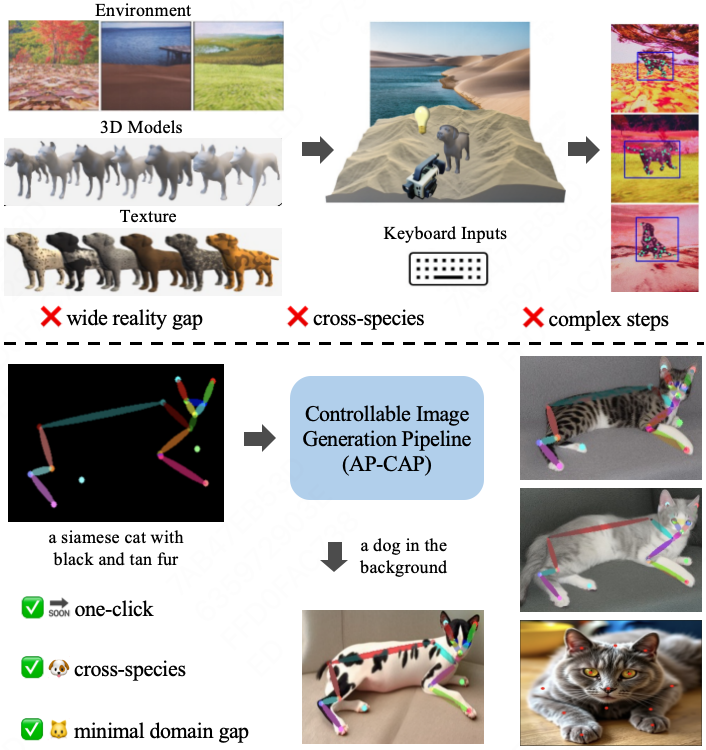} 
\end{center}
    \caption{Difference between traditional Animal Image Generation Paradigm and Ours. Top: Traditional 3D Modeling \& Color Rendering Pipeline. Bottom: Our End-to-End Controllable  Animal Image Generation Pipeline.}
\label{fig:intro}
\end{figure}

In response to its above diverse applications, recently, animal pose estimation has gained significant traction in the research community \cite{li2023scarcenet, chen2024grmpose, xu2022vitpose,zhang2022animal}. Current methods predominantly rely on supervised learning, leveraging annotated datasets to predict keypoints from visual data. While these approaches have demonstrated remarkable results, their effectiveness is fundamentally constrained by the availability of high-quality annotated data. The vast diversity of animal species, coupled with their complex morphologies, dynamic postures, and natural habitats, renders data collection and annotation both costly and challenging. Besides, existing animal pose datasets  remain limited in scale compared to human pose datasets. To address these challenges, various synthetic data generation techniques have been proposed \cite{mu2020learning, deane2021dynadog+, jiang2023spac}. These methods typically rely on creating 3D animal models, which are rendered into images with annotated keypoints. While this approach has improved data diversity to some extent, it remains limited by significant drawbacks. Generating diverse poses and realistic textures is computationally expensive, and the resulting images often suffer from a domain gap relative to real-world data, limiting their effectiveness in training robust animal pose estimation models.

Meanwhile, with the rapid development of generative artificial intelligence, controllable image generation, particularly diffusion models (e.g., Stable Diffusion) \cite{rombach2022high, zhang2023adding, luo2023latent, lu2024coarse}, has made realistic and diverse images generation increasingly feasible, without relying on 3D model reconstruction and rendering. This naturally raises an intriguing question: \empty{\textbf{can controllable image generation models, such as diffusion models, be harnessed to create diverse, high-quality synthetic datasets tailored specifically for training animal pose estimation models?}}

The answer is undoubtedly \textbf{`Yes'}. To this end, motivated by the above analysis, as shown in Fig.\ref{fig:intro}, we propose a novel Controllable Image Generation Pipeline, termed AP-CAP, for  high-quality annotated animal images synthesizing.  At the core of the proposed pipeline lies the Multi-Modal Animal Image Generation Model, which synthesizes images with precise, expected poses by leveraging a pretrained diffusion model. This model takes as input a seed image, target pose images, and text descriptions, enabling fine-grained control over the generated outputs. To further improve the quality and diversity of the synthesized data, we introduce three novel strategies: the Modality-Fusion-Based Animal Image Synthesis Strategy (MF-AISS), which fuses cross-modal features from text and images to generate visually diverse samples that are strictly aligned with the target poses; the Pose-Adjustment-Based Animal Image Synthesis Strategy (PA-AISS), which employs geometric transformations to enhance pose diversity by dynamically adjusting the limbs and torsos of input poses; and the Caption-Enhancement-Based Animal Image Synthesis Strategy (CE-AISS), which utilizes semantic understanding for text-guided generation control, producing samples with substantially varied poses while maintaining consistent semantic coherence. Together, these components form a comprehensive framework for synthesizing high-quality, diverse annotated data for animal pose estimation.

Using the proposed model and strategies, we synthesize the MPCH Dataset (Modality-Pose-Caption Hybrid), a large-scale hybrid dataset combining synthetic and real data. MPCH includes three subsets: the mammal pose dataset AP10k \cite{yu2021ap}, the multi-species dataset Animal-Pose \cite{cao2019cross}, and the diverse bird dataset Animal Kingdom-Birds \cite{ng2022animal}. Each subset supports intra-domain and cross-domain evaluations, with a 6:1 ratio of synthetic to real data for domain-specific benchmarks and diverse distribution tests.
Extensive experiments on various animal pose estimation models demonstrate that our synthesized data consistently enhances the performance of existing models in both intra-domain and cross-domain evaluations. Our contribution can be summarized as:
\begin{itemize}
    \item  We propose AP-CAP, a novel pipeline that generates annotated animal pose estimation data through controllable image generation, enabling precise pose and appearance synthesis.
\item  We design a diffusion-based generative model alongside three novel strategies: modality fusion, pose adjustment, and caption enhancement, to improve data quality and diversity.
\item  We create MPCH utilizing AP-CAP, the first large-scale hybrid dataset combining synthetic and real data, providing a comprehensive benchmark for animal pose estimation.
\end{itemize}

\section{Related Work}
\label{sec:related}
\subsection{Animal Pose Estimation Methods} 

Animal pose estimation involves identifying the spatial coordinates of keypoints on an animal's body from visual input, with methods spanning both 2D \cite{li2021synthetic, ye2024superanimal, li2023scarcenet} and 3D \cite{han2024multi, shooter2024digidogs, yang2024robust} domains. Despite recent progress, the scarcity of large-scale, high-quality datasets remains a major bottleneck, particularly for 2D pose estimation. Existing research primarily tackles this issue through (1) transfer learning from human pose data \cite{ye2024superanimal, xu2022vitpose, cao2019cross}, and (2) synthetic data generation to expand dataset diversity \cite{li2021synthetic, ye2024superanimal, li2023scarcenet}. However, these approaches face challenges such as domain gaps between synthetic and real-world data and the difficulty of generating accurate pseudo-labels. To address the limitations, this paper leverages a controllable image generation model to produce high-quality, realistically annotated synthetic data, aiming to enhance the performance of existing animal pose estimation models.

\subsection{Animal Data Synthesis Methods} 
Current methods for animal image synthesis primarily rely on 3D modeling and rendering pipelines. For example, CAD models serve as the foundation for generating synthetic animal images in \cite{mu2020learning}, where consistency standards refine pseudo-labels. Similarly, \cite{li2021synthetic} applies an iterative optimization strategy to improve pseudo-label accuracy. \cite{shooter2021sydog} uses the Unity3D engine to create synthetic canine datasets by adjusting dynamic environmental lighting and multi-view camera setups. In another approach, \cite{jiang2023spac} combines 3D animal models with realistic background images based on the ControlNet framework. While these methods expand available datasets, they encounter challenges such as complex, multi-stage modeling and rendering processes, which are computationally demanding. Additionally, a considerable domain gap often exists between synthetic data and real-world scenarios, limiting the effectiveness of these methods for tasks like animal pose estimation. In contrast, we use an end-to-end  controllable image generation model for image synthesis, which is more simple yet  effective.

\subsection{Controllable Image Generative Methods} 
The advent of diffusion models propels rapid advancements in image generation, resulting in the development of numerous diffusion-based methods \cite{brooks2023instructpix2pix, feng2022training, guo2023animatediff, liu2024towards, wu2023harnessing, jiang2023spac, ma2024deepcache} that demonstrate significant potential for high-quality image synthesis. Representative achievements in this field include DALLE-3 \cite{betker2023improving} and Stable Diffusion \cite{rombach2022high}, which set benchmarks for generative capabilities. Various extensions of diffusion models further enhance their flexibility and control. For example, ControlNet \cite{zhang2023adding} leverages multimodal conditional controls, such as pose, edges (Canny), and depth, to guide the generation process. Similarly, \cite{lu2024coarse} introduces a Coarse-to-Fine Latent Diffusion framework designed specifically for pose-guided person image synthesis tasks. Despite the impressive capabilities of these methods, few, if any, have been applied to synthesizing datasets for animal pose estimation in and end-to-end way; to the best of our knowledge, we are the first to explore this direction.

\section{Controllable Image Generation Pipeline}
\label{sec: methodology}
In this section, we introduce the controllable image generation pipeline designed for animal pose estimation data synthesis. The framework integrates three coordinated strategies—MF-AISS, PA-AISS, and CE-AISS—enabling single forward-pass inference to generate diverse data with varying poses and appearances. Our solution adopts an end-to-end single-stage training paradigm, reducing model complexity. Using this pipeline, we construct MPCH, a large-scale hybrid dataset that combines synthetic and real data, providing cross-domain, multi-level supervision. Next, we introduce the  architecture and the three strategies in detail. 

\begin{figure*}
\begin{center}
\includegraphics[width=\linewidth]{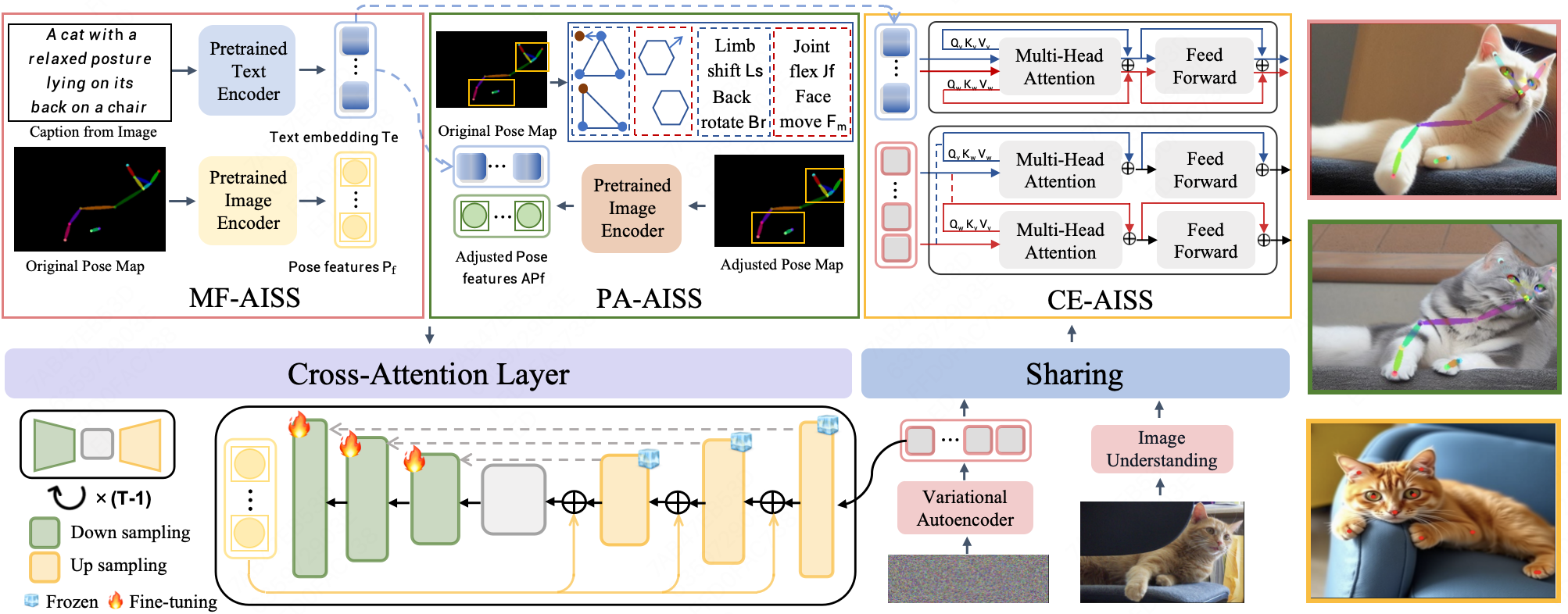} 
\end{center}
   \caption{Controllable Image Generation Pipeline, consisting of three strategies: Modality-Fusion-Based Animal Image Synthesis Strategy (MF-AISS), Pose-Adjustment-Based Animal Image Synthesis Strategy (PA-AISS), and Caption-Enhancement-Based Animal Image Synthesis Strategy(PA-AISS).}
\label{fig:framework}
\end{figure*}

\subsection{Architecture and Overview}
Fig.\ref{fig:framework} shows the architecture of our proposed method. The controllable image generation model is  built upon the latent diffusion model \cite{rombach2022high} with high-quality image generation capability, achieving robust synthesis through a single unified training phase. 
The training process consists of two key stages: (1) a Variational Autoencoder (VAE) \cite{esser2021taming} that establishes mappings between the raw-pixel space and low-dimensional latent space, and (2) a UNet-based \cite{ronneberger2015u} prediction model that utilizes text embeddings and pose features as conditional inputs to guide the denoising diffusion process for image generation.
To achieve enhanced control over texture synthesis, we introduce a Hybrid-Granularity Attention (HGA) module into the up-sampling blocks of the U-Net as described in \cite{lu2024coarse}.
Following the general idea of Denoising Diffusion Probabilistic Model (DDPM) \cite{ho2020denoising},
which formulates a forward diffusion process and a backward denoising process of T = 1000 steps.
The diffusion process progressively adds random Gaussian noise \(\epsilon \sim \mathcal{N}(0, I)\) to the initial latent \( z_0 \), mapping it into noisy latents \( z_t \) at different timesteps \( t \in [1, T] \):
\begin{equation}
z_t = \bar{\alpha}_t z_0 + \sqrt{1 - \bar{\alpha}_t},
\label{eq:zt}
\end{equation}
where \(\bar{\alpha}_1, \bar{\alpha}_2, \ldots, \bar{\alpha}_T\) are derived from a fixed variance schedule. The denoising process learns the UNet \(\epsilon_\theta(z_t, t, c, pf)\) to predict the noise and reverse this mapping, where \(c\) is the conditional text embedding output, where \(pf\) is the pose features. The optimization can be formulated as:
\begin{equation}
\label{eq:mse}
\mathcal{L}_{\text{mse}} = \mathbb{E}_{\bm{z}_0, \bm{c}, \epsilon, t} \left[ \| \epsilon - \epsilon_\theta(\bm{z}_t, t, \bm{c}, \bm{pf}) \|_2^2 \right].
\end{equation}

 During the inference stage, we take the input image and its original pose map as the baseline and use an image understanding interpreter (InternVL2-2B \cite{chen2024internvl}) to extract caption.

\subsection{Control Strategies}

In our method, to achieve broader pose diversity and improve the training of animal pose estimation models, we design three collaborative synthetic strategies: (1) MF-AISS, (2) PA-AISS, and (3) CE-AISS. These strategies generate diverse annotated data by varying poses and appearances in a controlled manner. Next, we introduce them in detail:

\noindent\textbf{MF-AISS:} This strategy processes the input text using the CLIP \cite{radford2021learning} Text Encoder to derive text embeddings \( T_s \), which are incorporated into cross-attention layers during both the down-sampling and up-sampling stages of the U-Net. Simultaneously, the input pose map is encoded by a lightweight image encoder based on ResNet blocks \cite{he2016deep}, producing pose features \( P_f \). These pose features are added to the output of each down-sampling block, following the approach in \cite{mou2024t2i}. The text embeddings \( T_s \) and pose features \( P_f \) act as collaborative conditioning signals, guiding the U-Net to generate images in a text-driven manner while adhering to the input pose constraints. This ensures visually diverse outputs with consistent poses, enhancing appearance generalization under fixed pose conditions.

\noindent\textbf{PA-AISS:} To address the challenges posed by unreasonable poses, which can reduce alignment between generated outputs and expected poses, PA-AISS introduces controlled variations to input poses while ensuring their rationality. This is achieved through four operations: (1) \textit{Face Move} \( F_m \), which adjusts the facial region's position relative to other body parts while respecting the symmetry and small distances of facial keypoints; (2) \textit{Limb Shift} \( L_s \) and \textit{Joint Flex} \( J_f \), which modify limb positions dynamically—applying global shifts for closely spaced limbs and random offsets (limited to half the inter-keypoint distance) for widely spaced limbs; and (3) \textit{Back Rotate} \( B_r \), which introduces small perturbations to the spine and neck keypoints to simulate natural backbone rotations. These refined poses are combined with the text embeddings \( T_s \) from MF-AISS and processed by the U-Net, generating data with both pose variations and diverse appearances to expand the dataset's coverage.

\noindent\textbf{CE-AISS:} This strategy employs an image understanding interpreter to generate semantic captions from input images and uses them as guidance during synthesis. CE-AISS reuses the text embeddings \( T_s \) from MF-AISS, ensuring consistent conditioning across strategies, while employing a text-to-image generation backbone adapted from the Flux framework \cite{black_forest_labs_2024}. The backbone integrates alternating layers of dual-stream and single-stream blocks, where the dual-stream architecture separately processes textual semantics and latent space features, enhancing alignment between captions and generated outputs. This approach synthesizes images that maintain semantic consistency with the original input scene while exhibiting diverse poses and appearances. By restructuring both pose and visual features, CE-AISS provides additional diversity crucial for training robust animal pose estimation models.

Building on the characteristic that all three generation strategies rely on text guidance, and informed by previous studies \cite{ni2022imaginarynet, witteveen2022investigating}, which demonstrate that diverse prompts effectively enhance image diversity, we design dual task-oriented prompt strategies to address the limitations of image diversity caused by overfitting in generative models. Specifically: (1) By inputting differentiated question instructions into InternVL2-2B, we dynamically generate diverse image descriptions, enriching the variety of textual guidance.
(2) Without altering the core subject of the generation target (e.g., the animal category), we randomly reorganize descriptive words associated with the same type of animal. This ensures that identical poses lead to images with varied appearance features across multiple inference tasks. During training, we employ a consistent and minimalist prompt template, such as "A [animal category] is in the background" to ensure simplicity while maintaining flexibility for diverse prompt synthesis during inference.

\section{MPCH Dataset}
Using the proposed method, we construct the MPCH (Modality-Pose-Caption Hybrid) dataset, a large-scale animal pose estimation dataset that combines real and synthetic data. This dataset is built through the proposed Controllable Image Generation Pipeline and consists of  three subsets using the following datasets as seeds: (1) AP10K \cite{yu2021ap}: A mammalian pose dataset comprising 10,015 high-quality images, spanning 23 families and 54 mammalian species, annotated with 17 keypoints.   (2)AnimalPose \cite{cao2019cross}: A dataset including 5,000 annotated images across 5 animal categories, each labeled with 17 keypoints. (3) Animal Kingdom-Birds \cite{ng2022animal}: A dataset with 8,524 annotated bird images from 189 bird species, each annotated with 23 keypoints.  

In MPCH, each subset is composed of in-domain and cross-domain components. For the in-domain setting, animal bounding boxes are cropped from the original images and processed using the three proposed strategies (MF-AISS, PA-AISS, CE-AISS). Each strategy generates two groups of annotated data with distinct prompts, which are combined with the original data to form a new training set at a 1:6 ratio of original to synthetic data. For the cross-domain setting, AP-10K and AnimalPose adhere to established cross-domain protocols from previous works \cite{yu2021ap, cao2019cross}, ensuring consistency with existing benchmarks. Meanwhile, Animal Kingdom-Birds employs a custom category partitioning scheme to evaluate generalization across fine-grained bird species.

Despite the high quality of the generated data, errors are inevitable, including pose misalignment during pose-controlled image generation and detection errors when the pose estimator analyzes the generated images. To mitigate the negative impact of these errors on the training process of pose estimation models, we introduce a filtering mechanism during training, implemented through the \( L_{\text{filter}} \) loss function, as defined in Equation \ref{eq:loss_keypoint}. This loss function screens the generated pose data by setting the loss of invalid samples to zero, preventing them from interfering with model training. Specifically, only keypoints with a loss below a predefined threshold \(\epsilon\) are considered in the total loss computation:

\begin{equation}
L_{\text{filter}} = \sum_{k=1}^{N} 
\begin{cases} 
\ell(\hat{y}_k, y_k), & \text{if } \ell(\hat{y}_k, y_k) \leq \epsilon \\
0,                    & \text{otherwise}
\end{cases}.
\label{eq:loss_keypoint}
\end{equation}
Here, \(\hat{y}_k\) represents the predicted value, and \(y_k\) denotes the ground truth for the \(k\)-th keypoint. \(\ell(\hat{y}_k, y_k)\) is the loss between \(\hat{y}_k\) and \(y_k\), \(N\) is the total number of keypoints, and \(\epsilon\) is the threshold value. This mechanism ensures that invalid generated data does not negatively impact the training process, improving the robustness and generalization performance of pose estimation models trained on the MPCH dataset.

\section{Experiment}
\subsection{Implementation Details}
\label{sec: experiment}
To validate the effectiveness of our method, we benchmark the MPCH dataset using mainstream pose estimation architectures. The results are compared against those obtained from existing state-of-the-art generative models, and we further evaluate the cross-domain generation strategy. Additionally, the proposed \( L_{\text{filter}} \) loss, as defined in Equation \ref{eq:loss_keypoint}, is integrated into the training process of the animal pose estimator to mitigate the impact of noisy generated data.  For all experiments, except those comparing different network architectures, HRNet-w32 \cite{sun2019deep} is used as the backbone for pose estimation. Training and testing are conducted on an NVIDIA Tesla V100 GPU with 16GB memory. Following the evaluation protocols in \cite{cao2019cross, yu2021ap}, we use the mean average precision (mAP) as the primary evaluation metric for the AP10K and AnimalPose datasets. For the Animal Kingdom-Birds dataset, we adopt the Percentage of Correct Keypoints (PCK@0.05) metric, in line with prior work \cite{ng2022animal}.

\subsection{Performance Evaluation}

\begin{figure}[t]
\begin{center}
\includegraphics[width=\linewidth]{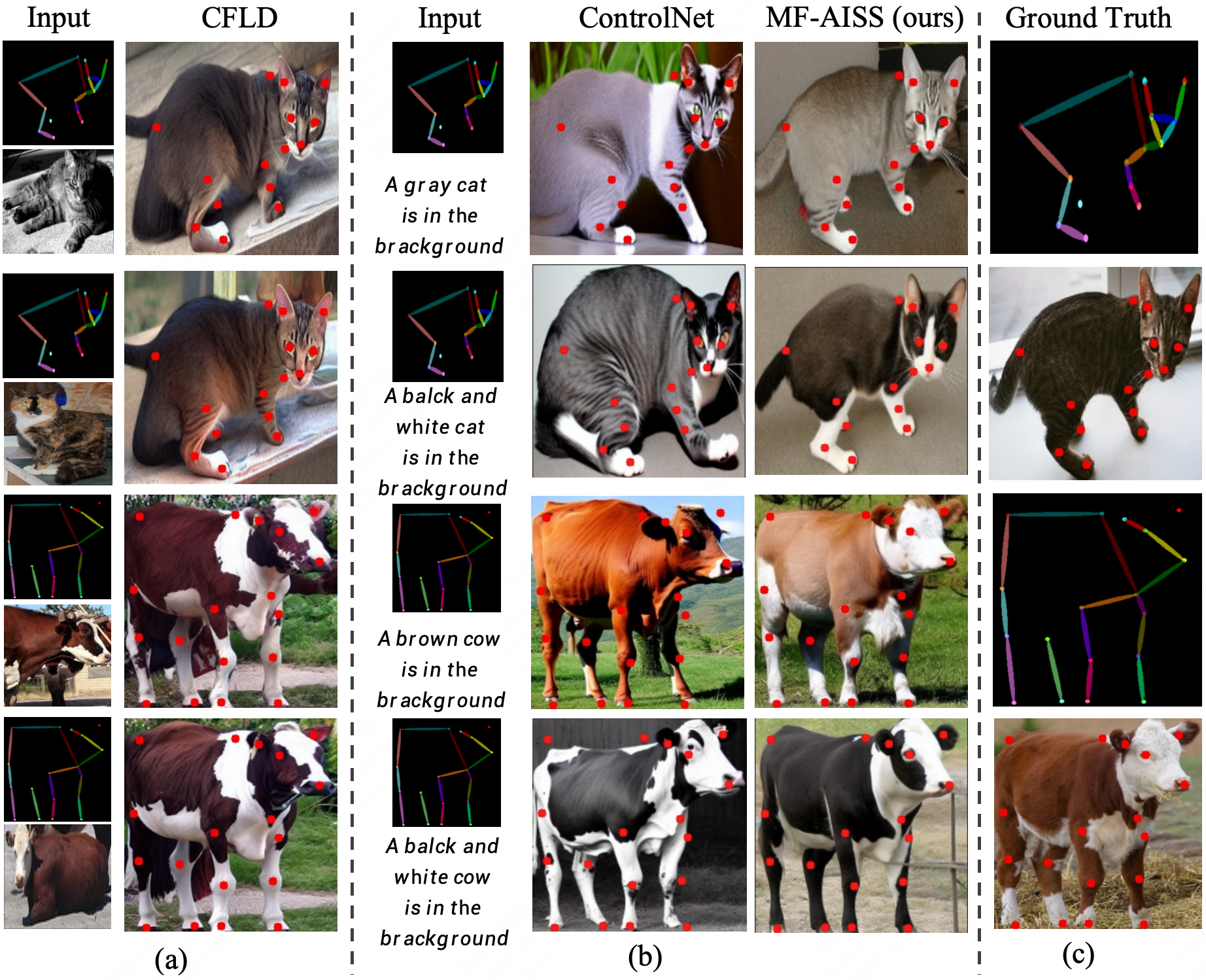} 
\end{center}
    \caption{(a) Synthetic data generation guided by input images and pose maps using CFLD \cite{lu2024coarse}. (b) Synthetic data generation controlled by text and pose maps using ControlNet \cite{zhang2023adding} and our method MF-AISS. (c) Real data.}
\label{fig:Quantitative}
\end{figure}

\begin{table*}[ht]
\centering
\begin{tabular}{@{}l *{6}{S[table-format=2.2]}@{}}
\noalign{\hrule height 1.2pt}
\multicolumn{6}{@{}l}{\textbf{Architecture Comparison}} \\ 
\noalign{\hrule height 0.8pt}
\multirow{2}{*}{Methods} & \multicolumn{2}{c}{AP10K (mAP)} & \multicolumn{2}{c}{Animal-Pose (mAP)} & \multicolumn{2}{c}{AK-Birds (PCK@0.05)} \\ 
\cmidrule(lr){2-3} \cmidrule(lr){4-5} \cmidrule(lr){6-7}
& {ORG} & {+AP-CAP} & {ORG} & {+AP-CAP} & {ORG} & {+AP-CAP} \\ 
\midrule
ResNet-101 \cite{he2016deep} & 69.89 & $\mathbf{72.92}$ & 69.02 & $\mathbf{72.87}$ & 75.39 & $\mathbf{77.14}$ \\
ViT-B \cite{xu2022vitpose}  & 72.60 & $\mathbf{74.45}$ & 72.28 & $\mathbf{73.78}$ & 77.09 & $\mathbf{78.03}$ \\
HRNet-w32 \cite{sun2019deep} & 73.36 & $\mathbf{76.20}$ & 73.50 & $\mathbf{76.59}$ & 76.93 & $\mathbf{78.94}$ \\
\noalign{\hrule height 1.2pt}
\multicolumn{6}{@{}l}{\textbf{Generation Strategy (HRNet-w32)}} \\ 
\noalign{\hrule height 0.8pt}
\multirow{2}{*}{Strategy} & \multicolumn{2}{c}{AP10K (mAP)} & \multicolumn{2}{c}{Animal-Pose (mAP)} & \multicolumn{2}{c}{AK-Birds (PCK@0.05)} \\ 
\cmidrule(lr){2-3} \cmidrule(lr){4-5} \cmidrule(lr){6-7}
& {Base} & {Improved} & {Base} & {Improved} & {Base} & {Improved} \\ 
\midrule
ORG & 73.36 & \text{--} & 73.50 & \text{--} & 76.20 & \text{--} \\
+ControlNet \cite{zhang2023adding} & 73.41 & 0.05 & 73.12 & {-}0.38 & 75.97 & {-}0.23 \\
+CFLD \cite{lu2024coarse} & 74.58 & 1.22 & 74.30 & 0.80 & 77.57 & 1.37 \\
\noalign{\hrule height 0.8pt}
+MF-AISS (Ours) & $\mathbf{75.36}$ & \textbf{+}$\mathbf{2.00}$ & $\mathbf{75.40}$ & \textbf{+}$\mathbf{1.90}$ & $\mathbf{78.51}$ & \textbf{+}$\mathbf{2.31}$ \\
\noalign{\hrule height 1.2pt}
\end{tabular}
\caption{Comprehensive comparison of architecture and generation strategies.}
\label{tab:hierarchical-comparison}
\end{table*}

To demonstrate the effectiveness of our proposed controllable image generation pipeline, we conduct comparative experiments on three mainstream pose estimation architectures: ResNet-101 \cite{ronneberger2015u}, ViT-B \cite{xu2022vitpose}, and HRNet-w32 \cite{sun2019deep}. Using the MPCH dataset, we construct a training set by combining generated data with real data at a 6:1 ratio under the intra-domain configuration. As shown in Table \ref{tab:hierarchical-comparison}, our proposed method consistently improves the performance of all tested pose estimation architectures, highlighting its robustness and adaptability. Notably, in the Animal-Pose dataset, HRNet-w32 achieves an improvement of 3.03 mAP! The improvements achieved by our method across diverse architectures underline its generalizability and the high-quality synthetic data it produces.

To further emphasize the superiority of our generation strategy, we compare our method against two advanced generative models: the general-purpose conditional control network, ControlNet \cite{zhang2023adding}, and the pose-guided image generation network, CFLD \cite{lu2024coarse}. In these experiments, we apply the MF-AISS strategy to augment the original dataset by twofold. The results in Table \ref{tab:hierarchical-comparison} clearly show that our method outperforms both ControlNet and CFLD, achieving the best performance metrics across all configurations. Specifically, on the AK-Birds dataset, our method achieves a 2.31 mAP improvement, surpassing CFLD by approximately 1 mAP! These results reinforce the effectiveness of our approach in producing high-quality and diverse pose-annotated data, which directly enhances the performance of downstream pose estimation tasks.

Fig. \ref{fig:Quantitative} provides a visual comparison of the generated outputs from our method and the two baseline models. While ControlNet \cite{zhang2023adding} demonstrates reasonable pose alignment through its general conditional control mechanism, it lacks the precision required for fine-grained keypoint alignment due to its limited focus on pose-specific constraints. According to Table \ref{tab:hierarchical-comparison}, as the number of keypoints in the dataset increases and the pose maps become more complex, ControlNet's performance degrades significantly. CFLD \cite{lu2024coarse}, as a pose-guided generation method, incorporates heatmap feature embeddings to improve keypoint consistency. However, its reliance on an image-guided paradigm restricts the diversity of generated samples, particularly in terms of species morphology, texture variations, and pose combinations. In contrast, our method effectively combines the strengths of fine-grained keypoint alignment and diverse appearance generation.

\subsection{Cross-Domain Pose Estimation}


\begin{table*}[ht]
\centering
\begin{tabular}{@{} l *{3}{S[table-format=2.2]} S[table-format=2.2] *{5}{S[table-format=2.2]} @{}} 
\noalign{\hrule height 1.2pt}
\multirowcell{3}{Method} & 
\multicolumn{3}{c}{AP10K (mAP)} & 
\multicolumn{1}{c}{\multirowcell{3}{AK-Birds\\(PCK@0.05)}} & 
\multicolumn{5}{c}{Animal-Pose (mAP)} \\ 
\cmidrule(lr){2-4} \cmidrule(lr){6-10}
& {Cervidae} & {Equidae} & {Hominidae} & & 
{Cat} & {Dog} & {Sheep} & {Cow} & {Horse} \\ 
\cmidrule(lr){2-4} \cmidrule(lr){6-10} 
& \multicolumn{3}{c}{} & & \multicolumn{5}{c}{} \\[-1.5ex]
\noalign{\hrule height 0.8pt}
\makecell[l]{\footnotesize WS-CDA+PPLO\\ \footnotesize \cite{cao2019cross}} & \text{--} & \text{--} & \text{--} & \text{--} & 42.3 & 41.0 & 54.7 & 57.3 & 53.1 \\
\makecell[l]{\footnotesize Baseline (ORG) \cite{yu2021ap}} & 66.12 & 50.76 & 3.23 & 54.48 & 52.87 & 57.79 & 63.44 & 64.09 & 59.02 \\
\makecell[l]{\footnotesize \textbf{MF-AISS (Ours)} \\ \footnotesize (ORG+proposed)} & $\mathbf{70.00}$ & $\mathbf{62.99}$ & $\mathbf{3.52}$ & $\mathbf{61.01}$ & $\mathbf{62.74}$ & $\mathbf{62.99}$ & $\mathbf{65.79}$ & $\mathbf{68.15}$ & $\mathbf{64.53}$ \\
\noalign{\hrule height 1.2pt}
\end{tabular}
\caption{Cross-dataset performance comparison of HRNet-w32.}
\label{tab:cross}
\end{table*}

\begin{figure*}
\begin{center}
\includegraphics[width=\linewidth]{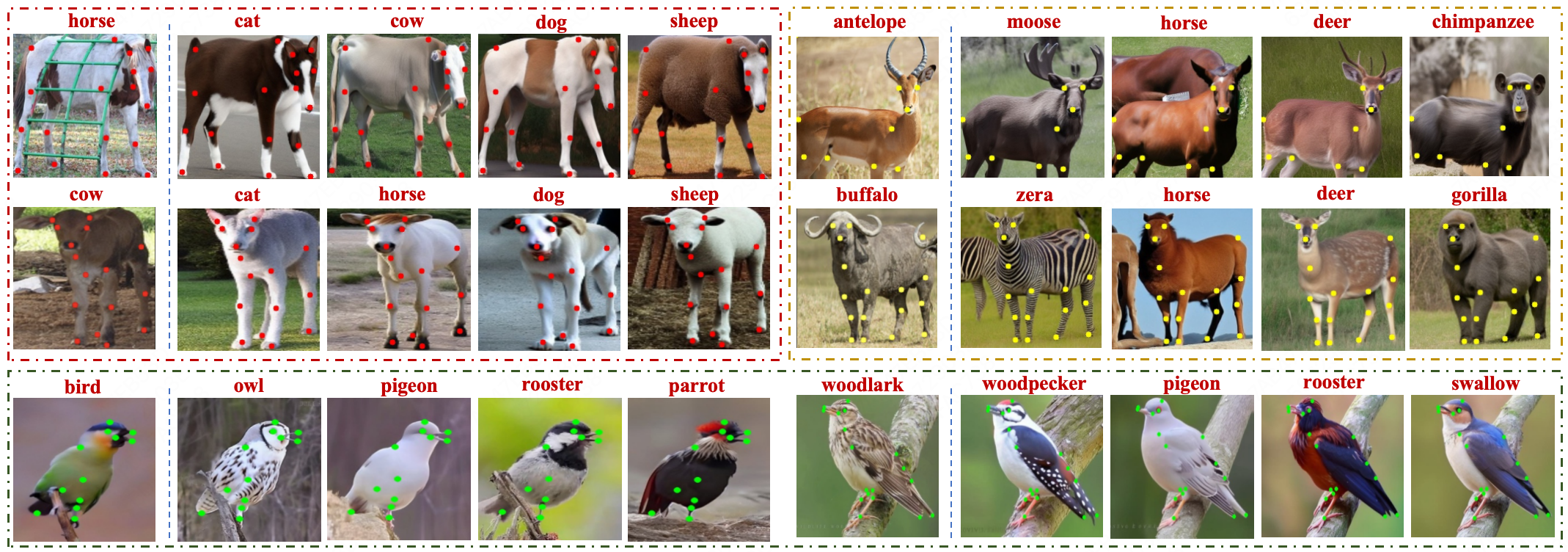} 
\end{center}
   \caption{Cross-domain data synthesis based on the MF-AISS strategy. Red dashed: AnimalPose; Yellow: AP10K; Green: Animal Kingdom-Birds. Column 1 shows real samples, with subsequent columns displaying generated results.}
\label{fig:cross}
\end{figure*}

In animal pose estimation research, the diversity of species encountered in practical applications far exceeds the coverage of existing annotated datasets. Annotating data for all animal species of interest is impractical, making cross-domain pose estimation a critical task. To address this challenge, we construct a cross-domain evaluation framework on our proposed MPCH dataset to systematically assess the effectiveness of our method. This framework allows us to evaluate how well our approach generalizes across species by synthesizing diverse cross-species pose samples. For example, by transferring annotated dog poses to unannotated species like foxes, we significantly enhance the cross-domain generalization ability of pose estimators.

Building on the above need for improved cross-domain generalization highlighted earlier, we evaluate our approach on several benchmark datasets under carefully designed cross-domain settings. For the AP10K and AnimalPose datasets, we adopt the established configurations from \cite{yu2021ap,cao2019cross} to ensure consistency with prior research. For the Animal Kingdom-Birds dataset, which comprises 189 categories, we design a custom cross-domain setup where the training set includes 158 categories (6,821 images) and the test set consists of 31 unseen categories (1,703 images). To further validate the effectiveness of our method, we use the MF-AISS framework to synthesize a target domain extension dataset, doubling the original data size. A balanced sampling strategy with a 1:2 ratio of source to target domain data ensures the generated samples complement the original training set. As shown in Table \ref{tab:cross}, our approach significantly improves generalization performance, with visual results in Fig. \ref{fig:cross} further demonstrating the diversity and realism of the generated cross-species pose samples. These findings confirm that our cross-domain generation strategy effectively bridges the gap posed by limited annotated data, enabling pose estimators to generalize across species with no direct annotations.

\subsection{Ablation Study}
\begin{figure*}[t]
\centering
\includegraphics[width=\linewidth]{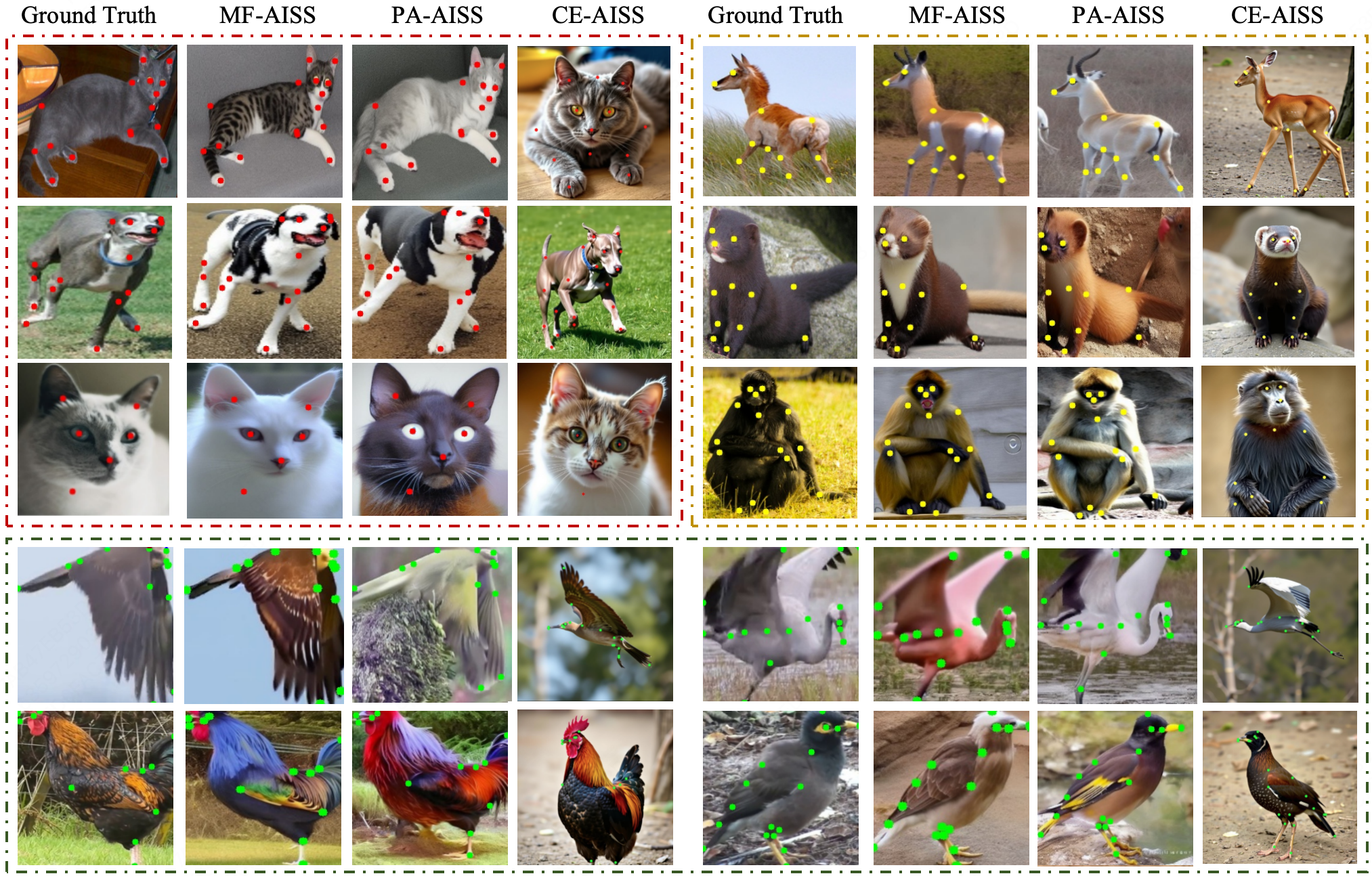}
\caption{Generated images of the MF-AISS, PA-AISS, and CE-AISS strategies.}
\label{fig:Ablation}
\end{figure*}

\begin{table}[!h]
\centering
\setlength{\tabcolsep}{2pt} 
\begin{tabular}{@{}ccc c S[table-format=2.2] S[table-format=2.2] S[table-format=2.2]@{}}
\noalign{\hrule height 1.2pt} 
\multicolumn{3}{c}{\makecell{Components}} & & \multicolumn{3}{c}{\makecell{Datasets}} \\
\cmidrule(lr){1-3} \cmidrule(l){5-7}
\makecell{MF-\\AISS} & \makecell{PA-\\AISS} & \makecell{CE-\\AISS} & & {AP10K} & {Animal-Pose} & {AK-Birds} \\
\midrule
\ding{55} & \ding{55} & \ding{55} & & 73.36 & 73.50 & 76.93 \\
\ding{51} & \ding{55} & \ding{55} & & 75.36 & 75.40 & 78.51 \\
\ding{51} & \ding{51} & \ding{55} & & 75.81 & 76.48 & 78.67 \\
\ding{51} & \ding{51} & \ding{51} & & $\mathbf{76.24}$ & $\mathbf{76.59}$ & $\mathbf{78.94}$ \\
\noalign{\hrule height 1.2pt} 
\end{tabular}
\caption{Component-wise Ablation Results of AP-CAP Framework on HRNet-w32: mAP of AP10K on the validation set, mAP of Animal-Pose and the PCK@0.05 of Animal Kingdom-Birds on the test set.}
\label{tab:ablation}
\end{table}

To evaluate the effectiveness of the proposed components, we conduct thorough ablation studies on the MPCH dataset, focusing on the contributions of MF-AISS, PA-AISS, and CE-AISS. The quantitative results are presented in Table \ref{tab:ablation}, while the corresponding visual results are shown in Fig. \ref{fig:Ablation}. Each component is analyzed in detail below.

\noindent\textbf{The Impact of MF-AISS:}  
As shown in Table \ref{tab:ablation}, this strategy significantly improves pose estimation performance across all three subsets by generating pose-consistent yet appearance-diversified data, achieving an average improvement of approximately 0.2 AP. MF-AISS preserves the original pose annotation topology (i.e., spatial keypoint distribution) while introducing diverse appearance variations, such as lighting conditions, environmental backgrounds, and biological surface textures. 
This enhances the generalization capability of models by creating a more robust feature representation space.From a visual perspective, the Controllable Image Generation Network excels at generating geometrically aligned synthetic images for various animal poses, including static (e.g., standing, sitting) and dynamic (e.g., walking, running) ones. Through prompt engineering, it generates cross-breed morphological features (e.g., feather gradients in Animal Kingdom-Birds) and intra-species phenotypic variations (e.g., Animal-Pose, third row), while also adapting to multi-species scenarios in AP10K. These results demonstrate the versatility of MF-AISS in improving both appearance diversity and pose alignment, making it a valuable tool for advancing cross-species pose estimation.

\noindent\textbf{The Impact of PA-AISS:}  
The PA-AISS strategy improves the performance of pose estimators by dynamically adjusting the original poses to generate new and diverse training data. As noted in prior studies, models trained solely on datasets with identical poses are prone to overfitting specific pose types, reducing their ability to generalize to unseen poses. PA-AISS addresses this limitation by introducing controlled pose variations, thereby enhancing the model's sensitivity to keypoint positions and improving its generalization capabilities.   
Visual analysis demonstrates that PA-AISS achieves balanced adaptation to key pose variations (e.g., limb shift, face move, etc.) while preserving anatomical plausibility, maintaining equilibrium between alignment fidelity and generation diversity.

\noindent\textbf{The Impact of CE-AISS:}  
CE-AISS cleverly leverages an advanced Diffusion Transformer-based framework to achieve high-fidelity reconstruction of image content guided by textual semantics. By strategically using original image captions as conditional input, CE-AISS ensures semantic consistency while fully reconstructing critical appearance textures (e.g., fur color and patterns) and pose topologies (e.g., joint angles and limb extensions). Moreover, keypoints on the generated images are automatically annotated using a pre-trained pose estimator, enabling the creation of an enhanced dataset with both geometric and phenotypic diversity. Experiments demonstrate that CE-AISS further improves pose estimation performance, as shown in Table \ref{tab:ablation}. Visual comparisons reveal significant differences in keypoint distributions between generated and original images, demonstrating geometric diversity with preserved semantic consistency. CE-AISS provides effective dataset augmentation that balances appearance/pose variation and semantic reliability.

\section{Conclusion}
\label{sec: conclusion}
In this paper, we propose a Controllable Image Generation Pipeline to advance end-to-end high-quality animal pose estimation data synthesis. The pipeline integrates three key strategies—MF-AISS, PA-AISS, and CE-AISS—to generate highly diverse and pose-consistent data. Leveraging this pipeline, we construct MPCH, the first large-scale hybrid dataset that combines synthetic and real data to enhance the performance of animal pose estimators.  Our experiments demonstrate the effectiveness of our approach in multiple aspects: (1) MPCH significantly boosts the performance of animal pose estimators by integrating high-quality synthetic data; (2) MPCH enhances the model’s generalization ability to unseen animal categories, addressing the challenge of limited annotated data problem of existing methods; and (3) the proposed method is versatile and performs robustly across different pose estimation frameworks. 

\clearpage 
{
    \small
    \bibliographystyle{ieeenat_fullname}
    \bibliography{arxiv}
}
{
\clearpage
\setcounter{page}{1}
\setcounter{table}{0}
\setcounter{figure}{0}
\setcounter{section}{7}
\renewcommand{\thetable}{\Alph{table}}
\renewcommand{\thefigure}{\Alph{figure}} 
\maketitlesupplementary
\section{Appendix}

In the appendix, we present additional visualization results:
Fig. \ref{fig:bu1} and Fig. \ref{fig:bu4} demonstrate the MPCH dataset constructed using our proposed AP-CAP pipeline.
Fig. \ref{fig:bu2} shows superior alignment of image keypoints compared to the state-of-the-art generation algorithm ControlNet \cite{zhang2023adding}.
Fig. \ref{fig:bu3} highlights our method’s capability to manipulate image appearance through flexible text control, outperforming the advanced CFLD \cite{lu2024coarse} approach in dataset diversity.

\begin{table}[htbp]
\centering
\caption{Performance Comparison on APT-36K Dataset with HRNet-W32}
\label{tab:ap36k_results}
\sisetup{table-format=2.2}  
\begin{tabular}{l *{4}{S}}  
\toprule
\multicolumn{1}{c}{Method} & 
\multicolumn{1}{c}{mAP (\%)} & 
\multicolumn{1}{c}{$\text{AP}_{50}$} & 
\multicolumn{1}{c}{$\text{AP}_{75}$} & 
\multicolumn{1}{c}{AR} \\
\midrule
ORG (Baseline)    & 63.93 & 89.30 & 68.25 & 68.33 \\
AP-CAP (Ours)     & \textbf{65.80} & \textbf{89.77} & \textbf{71.37} & \textbf{69.94} \\
\bottomrule
\end{tabular}
\vspace{0.3cm}
\small
\end{table}

\textbf{Zero-shot validation experiment.} To validate the generalization performance, we evaluate our method on the APT-36K \cite{yang2022apt} animal pose estimation and tracking benchmark without fine-tuning. As shown in Table \ref{tab:ap36k_results}, the model achieves a 1.87 mAP improvement, confirming that synthetic data enhances generalization ability. These results are further supported by the pose estimation visualization samples in the supplementary video (see attached archive), which illustrate the model’s precision in capturing complex animal poses.

\begin{figure*}[t]
\centering
\includegraphics[width=\linewidth]{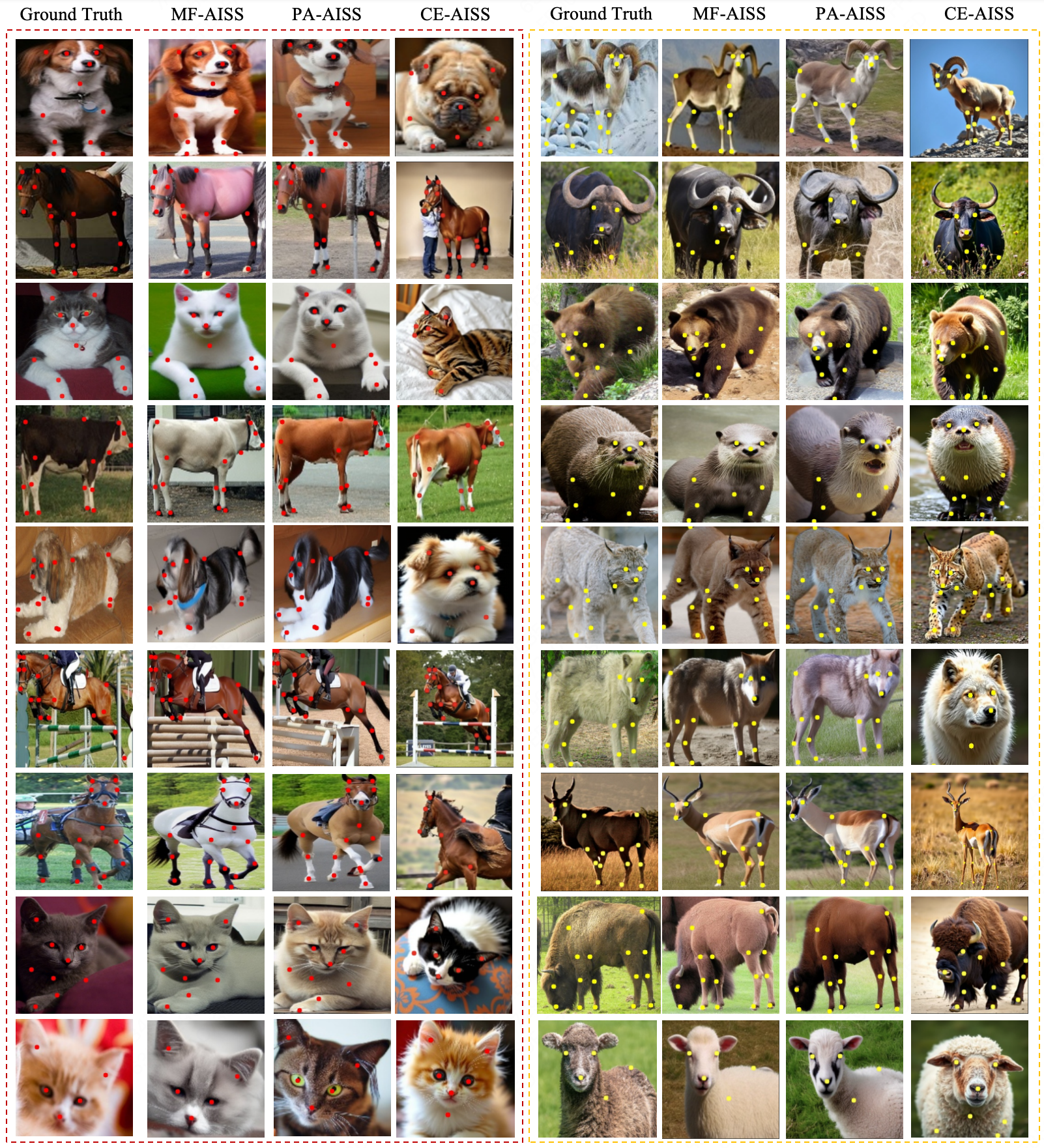}
\caption{The wildlife data presentation in the MPCH dataset is constructed using three strategies of the AP-CAP: MF-AISS, PA-AISS, and CE-AISS.}
\label{fig:bu1}
\end{figure*}
\appendix

\begin{figure*}[t]
\centering
\includegraphics[width=\linewidth]{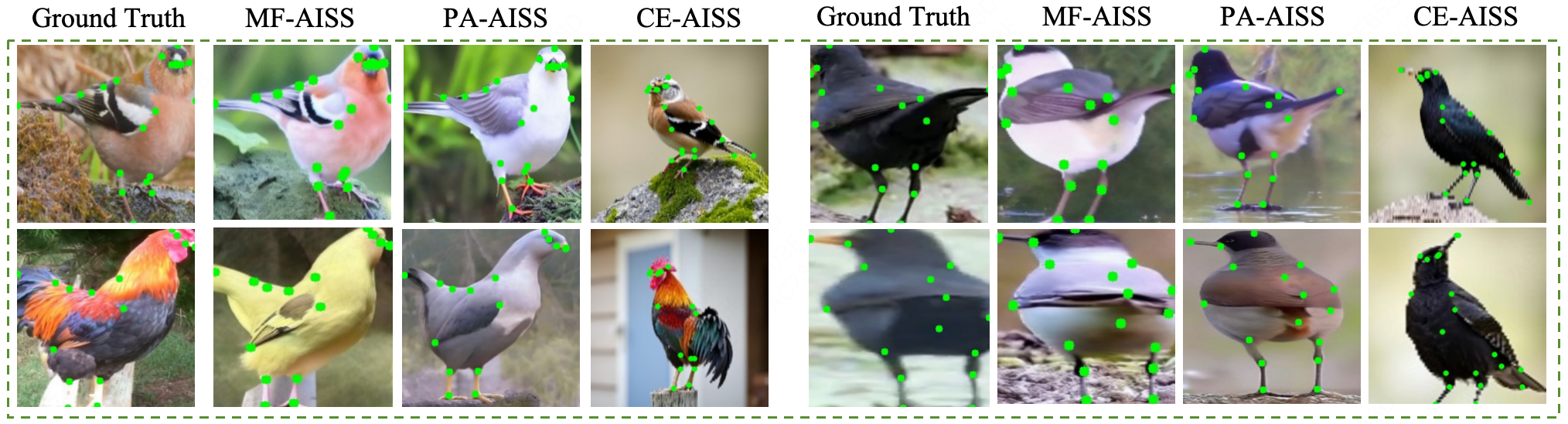}
\caption{A sample illustration of a subset of the bird data in the MPCH dataset.}
\label{fig:bu4}
\end{figure*}
\appendix

\begin{figure*}[t]
\centering
\includegraphics[width=\linewidth]{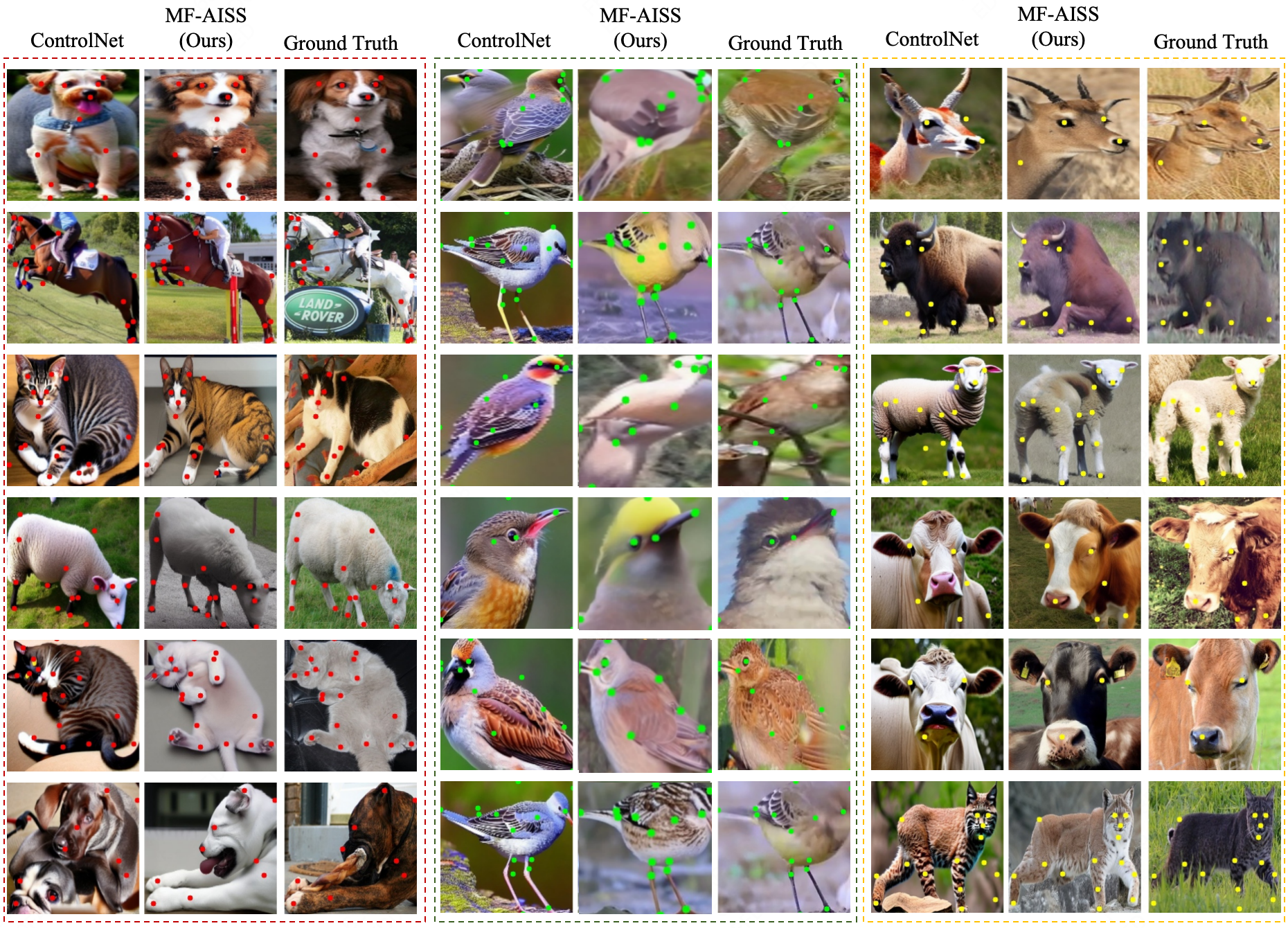}
\caption{The Visual Comparison of Synthetic Data between MF-AISS and ControlNet \cite{zhang2023adding}.}
\label{fig:bu2}
\end{figure*}
\appendix

\begin{figure*}[t]
\centering
\includegraphics[width=\linewidth]{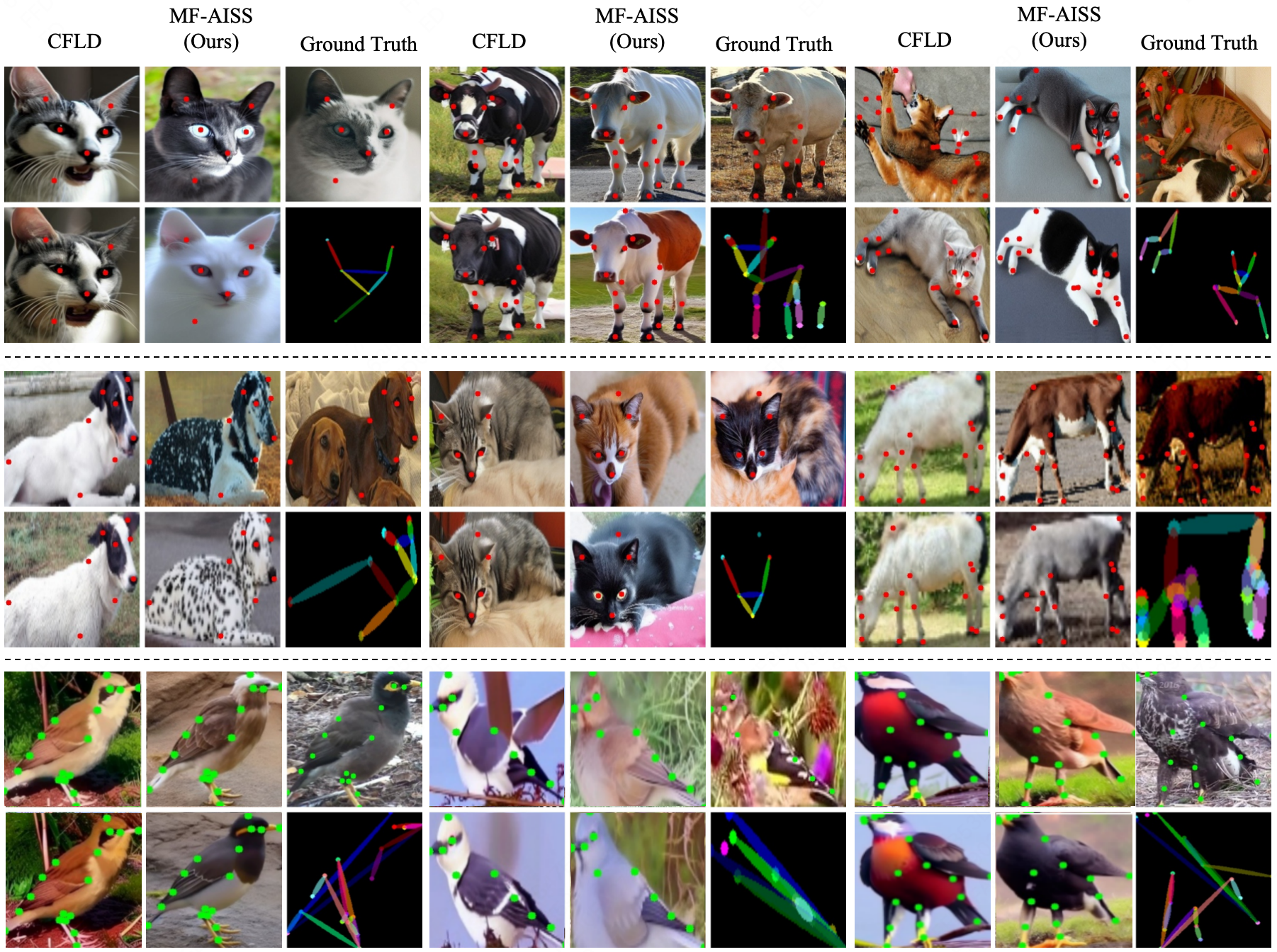}
\caption{The Visual Comparison of Synthetic Data between MF-AISS and CFLD \cite{lu2024coarse}.}
\label{fig:bu3}
\end{figure*}
\appendix

}
\end{document}